\documentclass{article}

\usepackage[final]{neurips_2023}

\usepackage[utf8]{inputenc} 
\usepackage[T1]{fontenc}    
\usepackage{hyperref}       
\usepackage{url}            
\usepackage{booktabs}       
\usepackage{amsfonts}       
\usepackage{bbding}       %
\usepackage{nicefrac}       
\usepackage{microtype}      
\usepackage[dvipsnames]{xcolor}         
\usepackage{graphicx}
\usepackage{diagbox}

\newcommand{\ours}{TableGPT}

\title{
\ours{}: Towards Unifying Tables, Nature Language and Commands into One GPT
}

\author{%
  Liangyu Zha$^{1,2}$ \quad Junlin Zhou$^{1,2}$ \quad Liyao Li$^{1,2}$ \quad Rui Wang$^{1,2}$ \quad Qingyi Huang$^{3}$ \\
  \textbf{\quad Saisai Yang$^{3}$ \quad Jing Yuan$^{3}$ \quad  Changbao Su$^{3}$ \quad Xiang Li$^{3}$ \quad Aofeng Su$^{3}$ \quad Tao Zhang$^{3}$} \\
  \textbf{} 
  \textbf{Chen Zhou$^{3}$ \quad Kaizhe Shou \quad Miao Wang \quad Wufang Zhu \quad Guoshan Lu \quad Chao Ye} \\
  \textbf{Yali Ye \quad Wentao Ye \quad Yiming Zhang \quad Xinglong Deng \quad Jie Xu }\\
  \textbf{
  Haobo Wang$^4$ \quad
    Gang Chen$^4$ \quad 
    Junbo Zhao$^4$\thanks{Correspondence to \texttt{j.zhao@zju.edu.cn}.}} \\
  \small{$^1$directional lead \quad $^2$joint first author \quad $^3$equal contribution \quad $^4$project lead}  \\
  \\
  Zhejiang University\\
}

\begin{document}

\maketitle
\setcitestyle{numbers,square}

\begin{abstract}
Tables are prevalent in real-world databases, requiring significant time and effort for humans to analyze and manipulate. 
The advancements in large language models (LLMs) have made it possible to interact with tables using natural language input, bringing this capability closer to reality.
In this paper, we present \ours{}, a unified fine-tuned framework that enables LLMs to understand and operate on tables using external functional commands.
It introduces the capability to seamlessly interact with tables, enabling a wide range of functionalities such as question answering, data manipulation (e.g., insert, delete, query, and modify operations), data visualization, analysis report generation, and automated prediction. \ours{} aims to provide convenience and accessibility to users by empowering them to effortlessly leverage tabular data.
At the core of \ours{} lies the novel concept of global tabular representations, which empowers LLMs to gain a comprehensive understanding of the entire table beyond meta-information. 
By jointly training LLMs on both table and text modalities, \ours{} achieves a deep understanding of tabular data and the ability to perform complex operations on tables through chain-of-command instructions. Importantly, \ours{} offers the advantage of being a self-contained system rather than relying on external API interfaces. Moreover, it supports efficient data process flow, query rejection (when appropriate) and private deployment, enabling faster domain data fine-tuning and ensuring data privacy, which enhances the framework's adaptability to specific use cases.

\end{abstract}

\section{Introduction}

The vast and intricate world of data is often encapsulated in tables, being a foundation for data-driven decision-making in a wide spectrum of applications, including financial analysis, supply chain management, and healthcare analytics.
It enables stakeholders to analyze trends, patterns, and relationships, leading to informed business decisions, process improvements, and resource optimization. 
For years, data scientists have struggled to process tables using complicated Excel formulas or handcrafted programming~\cite{li2022learning, lu2023catch}.
Consequently, there has been an urgent need to understand and interpret tabular data in a more efficient fashion. 

In the field of natural language processing, Generative Pre-trained Transformers (GPTs)~\cite{gpt1,gpt2,brown2020language,openai2023gpt4,chatgpt} or Large Language Models (LLMs)~\cite{chen2023phoenix, zeng2022glm, touvron2023llama, zhang2022opt} have \textbf{revolutionized} the paradigm of language data mining. 
Following this line of works, researchers have also explored large models for various modalities like vision~\cite{gong2023multimodal, kirillov2023segment}, and speech~\cite{huang2023audiogpt}. 
From a technical standpoint, their ability to generate human-like text has opened new vistas of possibilities for processing tabular data.
Nevertheless, it is non-trivial to directly employ the vanilla ChatGPT~\cite{chatgpt} model in the tabular area for two reasons:
(i)-\textbf{Global Table Understanding}: the GPTs are known to suffer from the limited token length and thus, they can not read a whole large table, making them hard to understand the global tabular information. 
(ii)-\textbf{Generalized to Tabular Domain: }
Second, their training processes are tailored for natural languages and thus, they are less generalizable when handling tabular data. 

\begin{table}[!t]
    \centering
    \small
    \caption{Comparisons with previous command-using LLMs for tabular data. (See details in Sec~\ref{sec:comparison})}
    \label{table:compare}
    \resizebox{\linewidth}{!}{
    \begin{tabular}{c|cccc}
    \toprule
        \diagbox{\textbf{Properties}}{\textbf{Methods}} & \textbf{ChatExcel~\cite{chatexcel}} & \textbf{SheetCopilot~\cite{li2023sheetcopilot}} & \textbf{Data-Copilot~\cite{zhang2023data}} & \textbf{\ours{}} (ours) \\ 
        \midrule
        Nature Language Operations & \color{OliveGreen}\Checkmark & \color{OliveGreen}\Checkmark & \color{OliveGreen}\Checkmark & \color{OliveGreen}\Checkmark \\ 
        Generalization to Arbitrary Tables & \color{OliveGreen}\Checkmark & \color{OliveGreen}\Checkmark & \color{YellowOrange}\XSolidBrush & \color{OliveGreen}\Checkmark \\ 
        Visualization & \color{YellowOrange}\XSolidBrush & \color{OliveGreen}\Checkmark & \color{OliveGreen}\Checkmark & \color{OliveGreen}\Checkmark \\ 
        Analysis \& Report & \color{YellowOrange}\XSolidBrush & \color{YellowOrange}\XSolidBrush & \color{OliveGreen}\Checkmark & \color{OliveGreen}\Checkmark \\ 
        Prediction & \color{YellowOrange}\XSolidBrush & \color{YellowOrange}\XSolidBrush & \color{OliveGreen}\Checkmark & \color{OliveGreen}\Checkmark \\ 
        Chain-of-command & \color{YellowOrange}\XSolidBrush & \color{YellowOrange}\XSolidBrush & \color{OliveGreen}\Checkmark & \color{OliveGreen}\Checkmark \\ 
        Base Model & Unknown & API & API & Fine-tuned \\ 
        Vague Input Rejection & \color{YellowOrange}\XSolidBrush & \color{YellowOrange}\XSolidBrush & \color{YellowOrange}\XSolidBrush & \color{OliveGreen}\Checkmark \\ 
        Private Deployment & \color{YellowOrange}\XSolidBrush & \color{YellowOrange}\XSolidBrush & \color{YellowOrange}\XSolidBrush & \color{OliveGreen}\Checkmark \\ 
        \bottomrule
    \end{tabular}}
\end{table}

There have been several works \cite{hu2023chatdb,zhong2017seq2sql,li2023nl2sql,li2023sheetcopilot} developed to integrate natural language for tabular data analysis. NL2SQL (Nature language to SQL)~\cite{hu2023chatdb,zhong2017seq2sql,li2023nl2sql} is a long-standing research topic that converts natural language to SQL commands that manipulate the relational database. Recently, SheetCopilot~\cite{li2023sheetcopilot} explored languages to VBA (Visual Basic for Applications, an embedded script language for Microsoft Excel) command such that benefit from a rich set of spreadsheet software functionalities. 
However, we found that both solutions demonstrate unsatisfactory performance. We speculate that these forms of programming code, which is fundamentally unstructured, adds another layer of complexity, making automated post-processing almost insurmountable. 

In this work, we develop \ours{} that pushes the boundaries of what is possible in data analysis empowered by LLM techniques, marking an important step forward in our pursuit of making data more accessible and understandable.
Our \ours{} framework unifies tables, natural language, and commands into a single GPT model, making data interpretation and manipulation more intuitive and user-friendly. 
By rethinking the interaction of tables, natural language, and commands, we integrate several core components into \ours{}:
\begin{itemize}
    \item \textbf{Global Table Representation: } We make the first attempt to develop a global representation learning paradigm for tables that encodes the whole table into one vector. By jointly training the LLM and a table encoder on vast amounts of text and table data, we equip the encoder to adequately capture the global information in the input table. This enables the LLM to perceive and understand the table data effectively, thereby providing a more global and enhanced comprehension of tables. 
    \item \textbf{Chain-of-Command: } We introduce this concept to emphasize the essential idea of a structured and hierarchical execution of tasks. Just like a well-coordinated organization where each directive is cascaded from a higher level to its lower counterpart, \ours{} follows a similar chain of commands, breaking down complex tasks into simpler ones and executing them step-by-step. Moreover, it fosters the ability to refuse ambiguous or inappropriate commands, much like an actual data scientist, instead of blindly following any potential erroneous instruction, thereby improving the interaction between humans and LLM systems in the field of data science. Our proposed command set is not only easier to control but also reduces the uncertainty that often accompanies traditional methods of handling table data.
    \item \textbf{Domain-aware Fine-Tuning: } To foster the ability to adapt to specific domains of tables and corresponding textual materials, domain-aware fine-tuning hinges on customizing training in a way that the model generates text embodying similar stylistic and logical elements found in a given domain, thereby augmenting its understanding of specific domain table data. To make this approach scalable and feasible, we have also developed a data processing pipeline that yields notable improvements with only a small amount of data, hence alleviating the resource-demanding aspect of training LLMs and supporting private deployment.
\end{itemize}

From a real-world production standpoint, the unstructured code outputted by NL2SQL poses significant challenges for preemptive checks and error corrections. Hence, we advocate for the use of structured command sequences, simplifying post-processing. 
Data-Copilot~\cite{zhang2023data} also embraces this command-based approach with self-instruct~\cite{wang2022self}, but its reliance on API-called native LLMs to comprehend tabular data's processing and analysis logic directly presents limitations. Given the intrinsic data variability and task-specificity of tabular data, we believe an effective product should be custom-built for tabular data while maintaining general applicability to broader downstream tasks. This conviction underscores the imperative of introducing a LLM specifically pre-trained for tabular data.

To sum up, this work presents a pioneering \ours{} framework, which is a unified, well-fledged holistic solution, enabling efficient tabular data processing, analysis and visualization, driven all by natural languages.
We summarize several important advantages of \ours{} as follows:
\begin{itemize}
    \item \textbf{Language-driven EDA: } \ours{} understands user intent from natural language, dissects the desired actions, and executes external commands on the table. It subsequently returns the processed results in both tabular and textual explanations to the user. This novel approach simplifies the way users engage with table data, bringing an intuitive instantiation to Exploratory Data Analysis (EDA). 
    \item \textbf{Unified  Cross-modal Framework: } Innovatively, we devise a global table encoder for understanding the whole table. \ours{} is able to fully understand the user query, metaknowledge, and whole tabular data, which leads to much more reliable execution commands for table manipulation. 
    \item \textbf{Generalization and Privacy: } By domain-aware fine-tuning, our \ours{} can better handle data variability of tables and generalize to different domains. Further, our framework supports private deployment, offering robust data privacy protections. This aspect is critical in the modern age where data privacy and protection are just paramount.
\end{itemize}

\begin{figure*}[!t]
    \centering
    \includegraphics[scale=0.4]{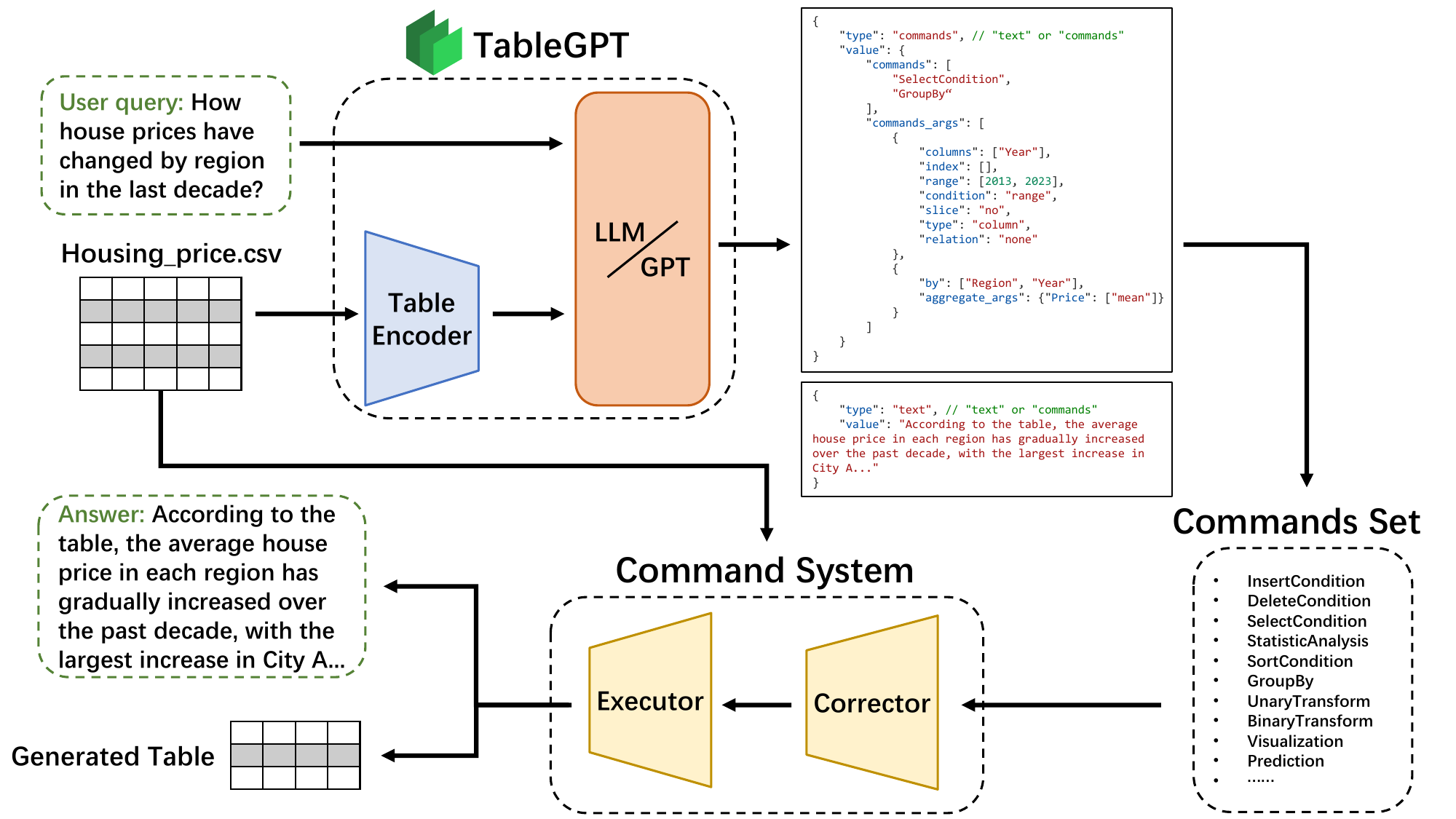}
    \caption{An architecture of \ours{} framework.}
    \label{fig:archi}
\end{figure*}

\section{\ours{}}
\subsection{Model Design}
The development of \ours{} begins with the foundation provided by pre-trained LLMs. The advancements in the field of natural language processing have led to the development of a number of exceptional open-source LLMs, such as LLaMa~\cite{touvron2023llama}, Phoenix~\cite{chen2023phoenix}, ChatGLM~\cite{zeng2022glm}, Ziya~\cite{Fengshenbang-LM}, and Baichuan~\cite{baichuan}.
In designing \ours{}, we opted to use Phoenix~\cite{chen2023phoenix} with 7B parameters as our base model for fine-tuning, owing to its excellent capabilities in handling both Chinese and English languages. This choice is not, however, exclusive. Our model design supports adaptation with other LLMs, providing versatility and flexibility in its implementation.

What sets \ours{} apart from its predecessors~\cite{chatexcel,li2023sheetcopilot,zhang2023data} is the novel approach to its fine-tuning process. We performed the fine-tuning on a vast corpus, comprising 2T tokens of textual data and 0.3M tables. This corpus offers a diverse landscape for the model to learn from, including but not limited to user query-command sequence pairs and publicly available domain-specific data for table analysis reports. 

The overall architecture of \ours{} is shown in Figure~\ref{fig:archi}. When a user inputs a table and a query, these are received by \ours{}, which consists of a table encoder and an LLM.
The table encoder serves to extract vector representations from the input table. These representations, coupled with the text query, are then fed into the LLM for inference. The LLM discerns the user's query intent and generates an output that includes both a command sequence and a textual reply.
The command sequence undergoes error correction in the command system's corrector before it is fed into the executor for execution. The final output, provided to the user, includes the manipulated table and a textual reply. This streamlined process delivers efficient, reliable responses to table data queries, enhancing user experience and simplifying data analysis.

\subsection{Global Representation of Table}
The rapid development of large language models (LLMs) has seen them interfacing with a multitude of modalities such as vision, and audio. For instance, the integration of vision and LLMs has led to models like CLIP~\cite{radford2021clip} (Contrastive Language–Image Pretraining) from OpenAI that connects images and text through shared latent space. The combination of audio and LLMs gave rise to models like Wave2Vec~\cite{baevski2020wav2vec} and Tacotron~\cite{wang2017tacotron} that employ the representation of audio in the form of spectrograms to generate or understand speech.

Despite these advancements, the exploration of LLMs interfacing with tabular data remains limited. The question of how to enable LLMs to comprehend and interpret tables is essential. Some studies have attempted to convert sample rows of table data directly into a sentence-like text description~\cite{hegselmann2023tabllm}, while others have attempted to artificially define a global representation of table data through the template-based extraction of column names, industry background, and other metadata schema~\cite{zhang2023data}. However, these approaches only extract partial information from table data for LLMs, consequently overlooking the global information and industry background inherent in the data.

Notably, for the tables, it is required to embed the whole table into one single vector, instead of generating sample-wise embedding. This can be non-trivial and challenging because, unlike images, videos, and audio, table data is inherently a highly abstract structured data type. 
Furthermore, it possesses a dual permutation invariance structure where shuffling rows or columns does not affect the information contained within the table, a distinct contrast to images and audio, which carry inductive bias in adjacent positions or sequences. Moreover, tables from different domains vary in size and format, such as having different numbers of discrete and continuous columns, making it challenging to extract features from diverse tables using a unified neural network architecture~\cite{ye2023ctbert}. 

Yet, it remains an open problem to effectively extract global representations from tables for LLMs to achieve comprehensive table understanding. To this end, we present a Cascaded Table Encoder that jointly extracts knowledge from metadata and whole numerical entries. 

\paragraph{Cascaded Table Encoder. } 
Consider the approach of an experienced data scientist encountering a table. They typically examine the structure of the table data, such as the table headers and distribution of feature columns, to understand the meaning of different cells based on their position, without focusing too much on the numeric information of each cell.
Following this biologically plausible approach, we propose a novel cascading table encoder. It divides the information in the table data into two main parts. The first part learns the metadata representation of the table, such as schema, industry background, and the meanings of column names, which can help LLMs understand the global information of the table structure. The second part learns the numerical information representation of the table, such as the distribution and trends of values in different columns, helping LLMs understand the global information of the table numbers like human experts.

We consider the rows and columns of the table as elements of a set and learn the overall representation of the entire set. We use a modified set transformer~\cite{lee2019settrans} as the backbone of the table encoder. The set transformer~\cite{lee2019settrans}, originally designed for dealing with permutation invariant problems, aligns well with the inherent structure of tabular data. We enhance it with an attention mechanism~\cite{vaswani2017attention} that can capture the interdependencies between different rows or columns of the table, enabling the model to understand the relations between different parts of the table data. 

This encoder is pre-trained on ten thousand table datasets using a masked table modeling approach, similar to the masked language modeling used in BERT~\cite{devlin2019bert} but adapted to tabular data. The learned table representation not only can be used for table understanding but also can enhance the predictive performance of downstream classifiers.

Our proposed method presents a significant step forward in the integration of tables, natural language, and commands into LLMs. It provides a comprehensive approach for extracting global representations from tables and enables LLMs to understand and manipulate.
 
\subsection{Chain-of-Command}
In recognition of the fact that Large Language Models (LLMs) like GPT can struggle with numerical reasoning, prone to computational errors and hallucinations~\cite{imani2023mathprompter}, our approach does not require them to operate and calculate within the tables in their latent space. Instead, we provide a series of pre-packaged function commands for LLMs to call upon. LLMs, understanding the global representation of the table and user input, generate a sequence of commands for the backend system to execute, resulting in a modified table. Compared to the SQL statements generated by text2SQL~\cite{hu2023chatdb,zhong2017seq2sql,li2023nl2sql}, these command sequences are more easily examined and error-located by the backend parsing system, while SQL statements can be challenging to diagnose and correct for specific errors.

However, user queries are often vague and complex, and we can only encapsulate and provide some basic table operation commands. Teaching the LLM to deconstruct complex and vague queries is crucial. For example, a user's query for a specified object column could be a synonym or translation of a column in the original table, or the user may only have a vague intent and cannot express the demand clearly.

The Chain-of-thought~\cite{kojima2022cot1,wei2022cot2} approach emphasizes breaking down complex reasoning into a series of intermediate steps. We introduce the concept of Chain-of-command (CoC), an approach that enhances the chain-of-thought by providing a mechanism for step-by-step instructions associated with these intermediate steps. For instance, when a user asks, "Show me the five movies with the highest profit margin," the LLM first checks if a profit margin column exists in the table. If not, it generates arithmetic instructions to calculate the profit margin using box office and cost data; next, it executes instructions to sort by profit margin in descending order and slice to select the top five movies. When user queries are too vague, like "Give me some numbers," the LLM might struggle to decompose and could refuse execution, instead, it would ask the user for more specific intent.

The aim of the Chain-of-command is to enhance LLM's reasoning capabilities and robustness when operating table data. This approach involves translating user inputs into a sequence of intermediate command operations, enabling LLMs to manipulate tables more accurately and efficiently symbolically. The ability to manipulate symbolic instructions is particularly valuable for real-world applications involving complex and accurate interactions with historical data, such as record-keeping and data analysis in management environments.

To enhance the performance and stability of our approach, we constructed a substantial dataset of command chain instructions while fine-tuning LLMs to adapt to commands, and employed contextual learning to provide prompts for multiple steps in the command chain sequence. A strong and accurate command chain process allows LLMs to better reason about table data and handle more complex scenarios.

The Chain-of-command approach has three main advantages. First, it enables LLMs to execute complex table instructions accurately, thereby enhancing their multi-hop reasoning capabilities for table operations. Second, by breaking down complex operations into a series of intermediate table operations, the chain-of-command method enhances the LLM's ability to handle complex multi-table interactions. Lastly, it enables LLMs to refuse overly vague instructions and ask users for more specific intent. This approach allows LLMs to handle edge cases and unexpected scenarios better, making it a promising method for real-world applications.

\subsection{Domain Data Processing Pipeline}
Despite the broad knowledge and dialogue capabilities of large language models (LLMs) due to extensive pre-training on a diverse corpus, their performance often falls short in addressing the nuanced language styles and logic of specific industries. This is primarily due to the lack of exposure to proprietary domain data during their training phase. To mitigate this issue, we have developed an efficient domain data processing pipeline~\cite{chen2023maybe, ye2023assessing}.

Motivated by the goal to streamline the fine-tuning process of LLMs with minimal computational overhead and accelerated model iteration, our pipeline is designed to harness the power of active learning~\cite{ren2021acsurvey}. Through this, we curate a carefully selected set of fine-tuning examples from the domain data, allowing LLMs to achieve superior fine-tuning results with a reduced number of examples. This strategic utilization of resources expedites the model's learning process, thereby speeding up its iteration.

Additionally, we have fortified the document retrieval capabilities of LLMs. We utilize technologies like vector databases~\cite{wang2021vecdb} and LangChain~\cite{langchain} to facilitate the retrieval of pertinent information from a plethora of proprietary documents, further enriching the context that LLMs learn from. 

In essence, our pipeline serves as a catalyst for the rapid and cost-effective adaptation of LLMs to the data needs of various specific industries. This pipeline not only addresses the challenges of industry-specific language styles and logic but also empowers LLMs to handle commands that interact with tables, integrating the realms of natural language, tables, and commands.

\section{Evaluation}

\subsection{Commands supported by \ours{}}
To unleash the power of \ours{}, we have designed and supported a rich set of commands.
Firstly, \ours{} enables natural language interaction with tables, empowering users to intuitively query, filter, sort, and aggregate data using everyday language. It also facilitates tasks such as data visualization and report generation, enhancing the interpretability and presentation of tabular information. Lastly, \ours{} facilitates automated decision-making processes, empowering users to make predictions, forecast trends, and estimate outcomes using table data and natural language instructions.

Note that when the intent of the user query is too vague, \ours{} will reject to generate commands and instead ask the user for more detailed intent. This is one of the benefits of chain-of-command, the ability to think about the rationality of commands like a human expert, rather than a rigid command translator.

\subsection{Comparison with previous command-using LLMs}~\label{sec:comparison}
Several existing solutions attempt to combine tables and language models, such as ChatExcel~\cite{chatexcel}, SheetCopilot~\cite{li2023sheetcopilot}, and Data-Copilot~\cite{zhang2023data}. These approaches typically rely on using prompts to invoke pre-defined external commands through inference API of LLMs, such as OpenAI API\footnote{https://openai.com/blog/openai-api}. In contrast, \ours{} takes a different approach by fine-tuning LLM specifically for table-related tasks. This key distinction allows us to harness the inherent capabilities of the LLM architecture while tailoring it to excel in table processing tasks.
A detailed comparison of \ours{} with the previous command-using LLMs is shown in Table~\ref{table:compare}.

\subsection{Case Study}
\begin{figure*}[!h]
    \centering
    \includegraphics[scale=0.34]{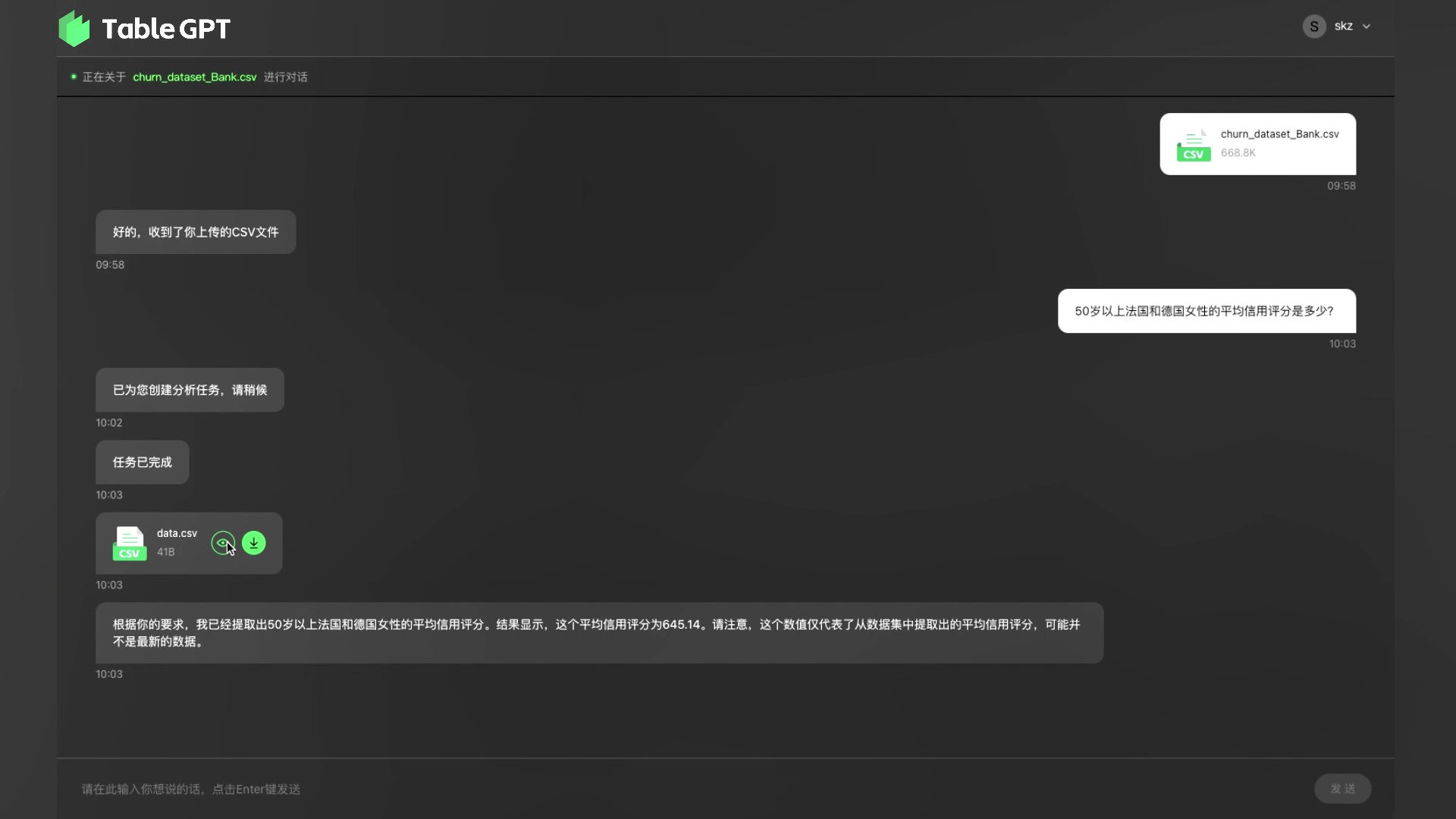}
    \caption{Cases of \ours{}.}
    \label{fig:1}
\end{figure*}

\begin{figure*}[!h]
    \centering
    \includegraphics[scale=0.34]{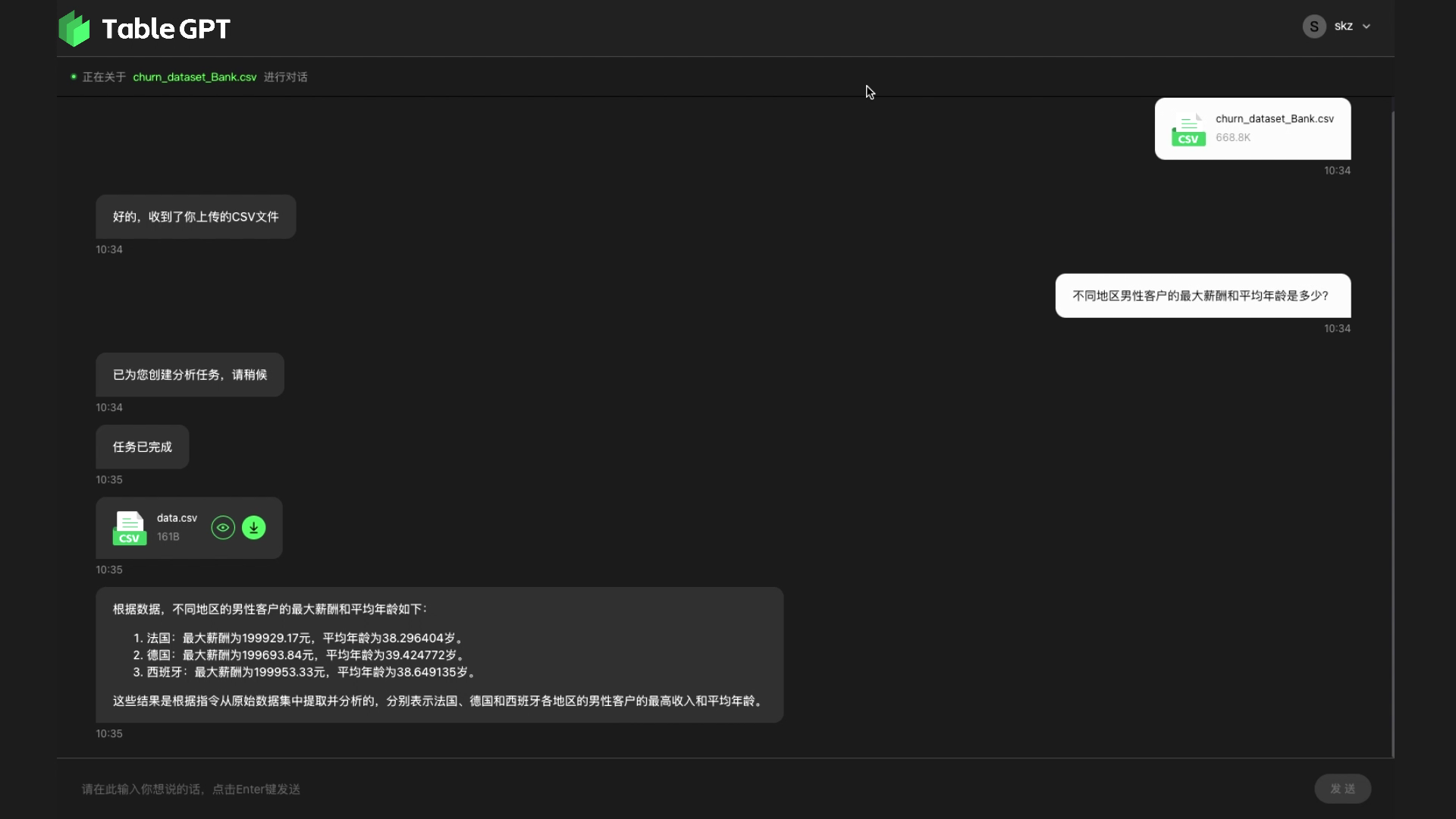}
    \caption{Cases of \ours{}.}
    \label{fig:2}
\end{figure*}

\begin{figure*}[!h]
    \centering
    \includegraphics[scale=0.34]{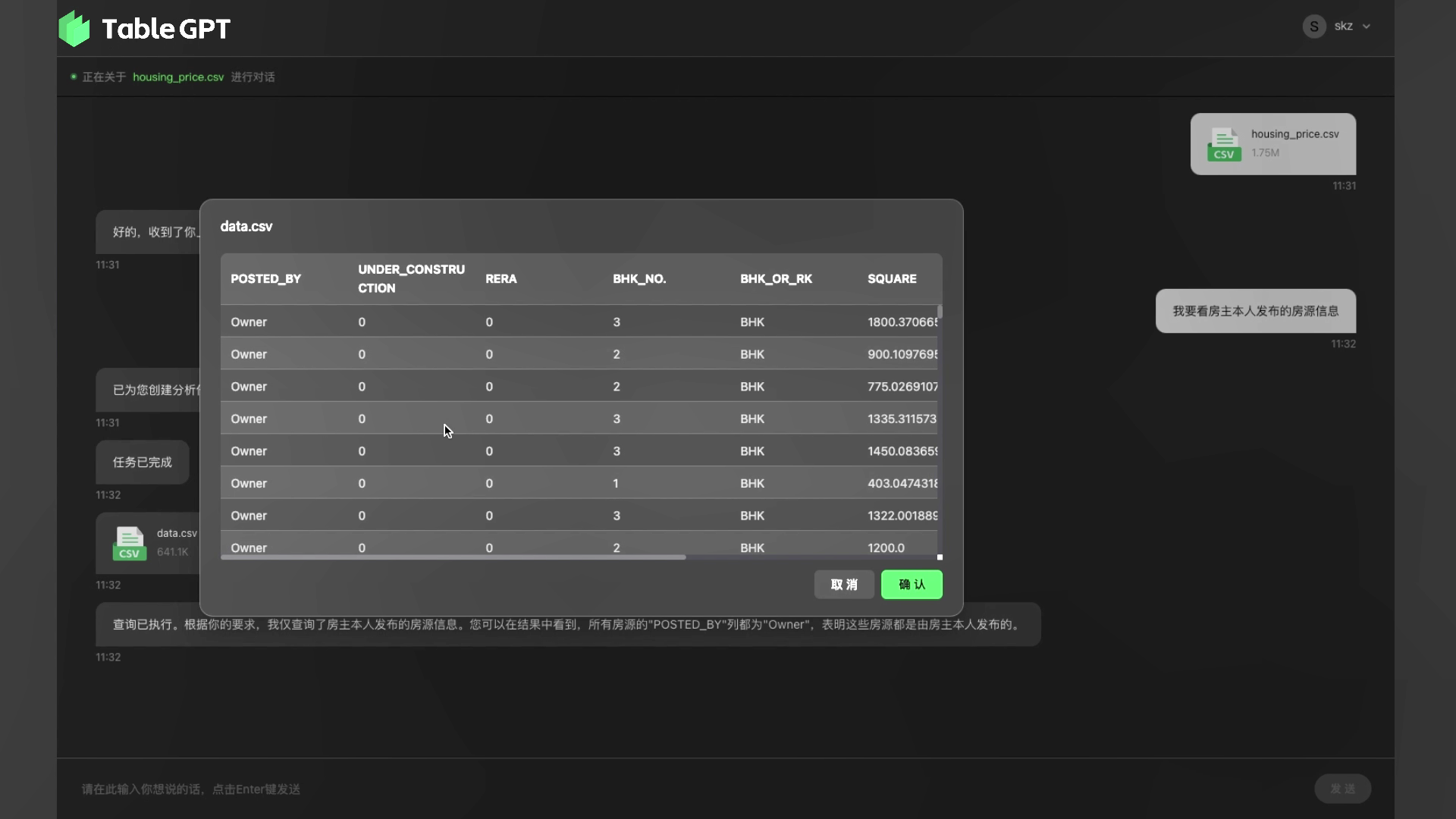}
    \caption{Cases of \ours{}.}
    \label{fig:3}
\end{figure*}

\begin{figure*}[!h]
    \centering
    \includegraphics[scale=0.34]{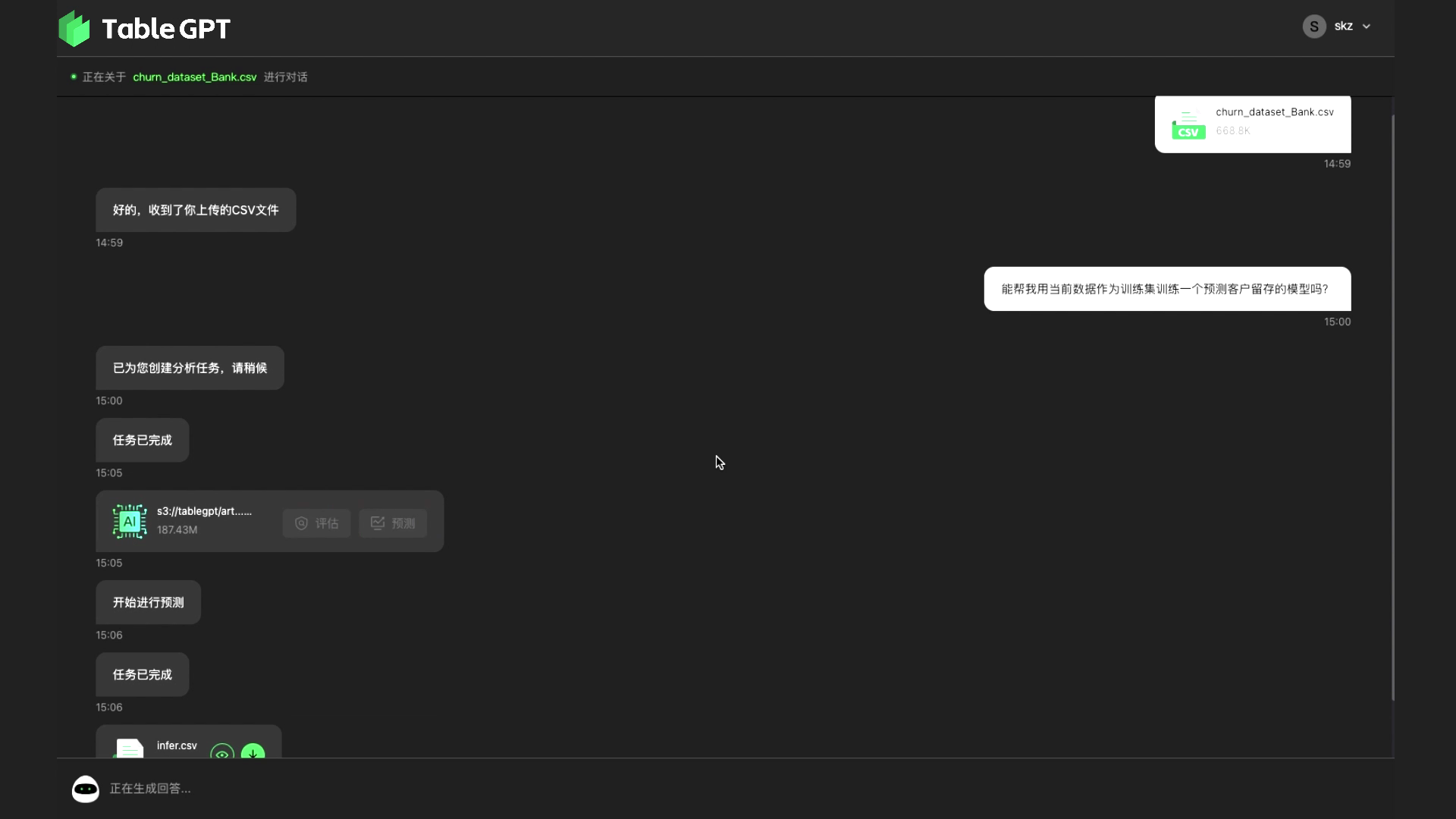}
    \caption{Cases of \ours{}.}
    \label{fig:4}
\end{figure*}

\begin{figure*}[!h]
    \centering
    \includegraphics[scale=0.34]{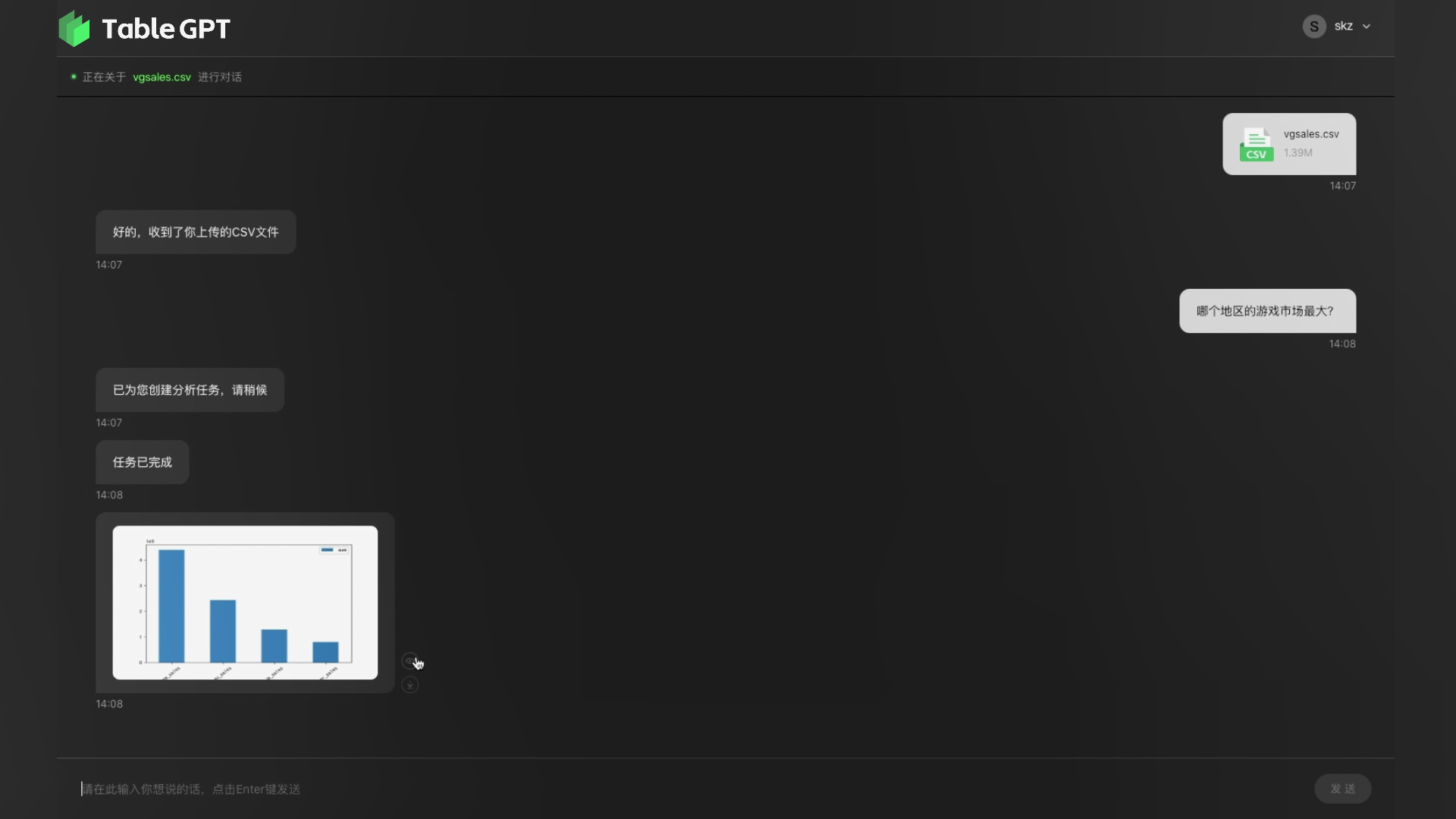}
    \caption{Cases of \ours{}.}
    \label{fig:5}
\end{figure*}

\begin{figure*}[!h]
    \centering
    \includegraphics[scale=0.34]{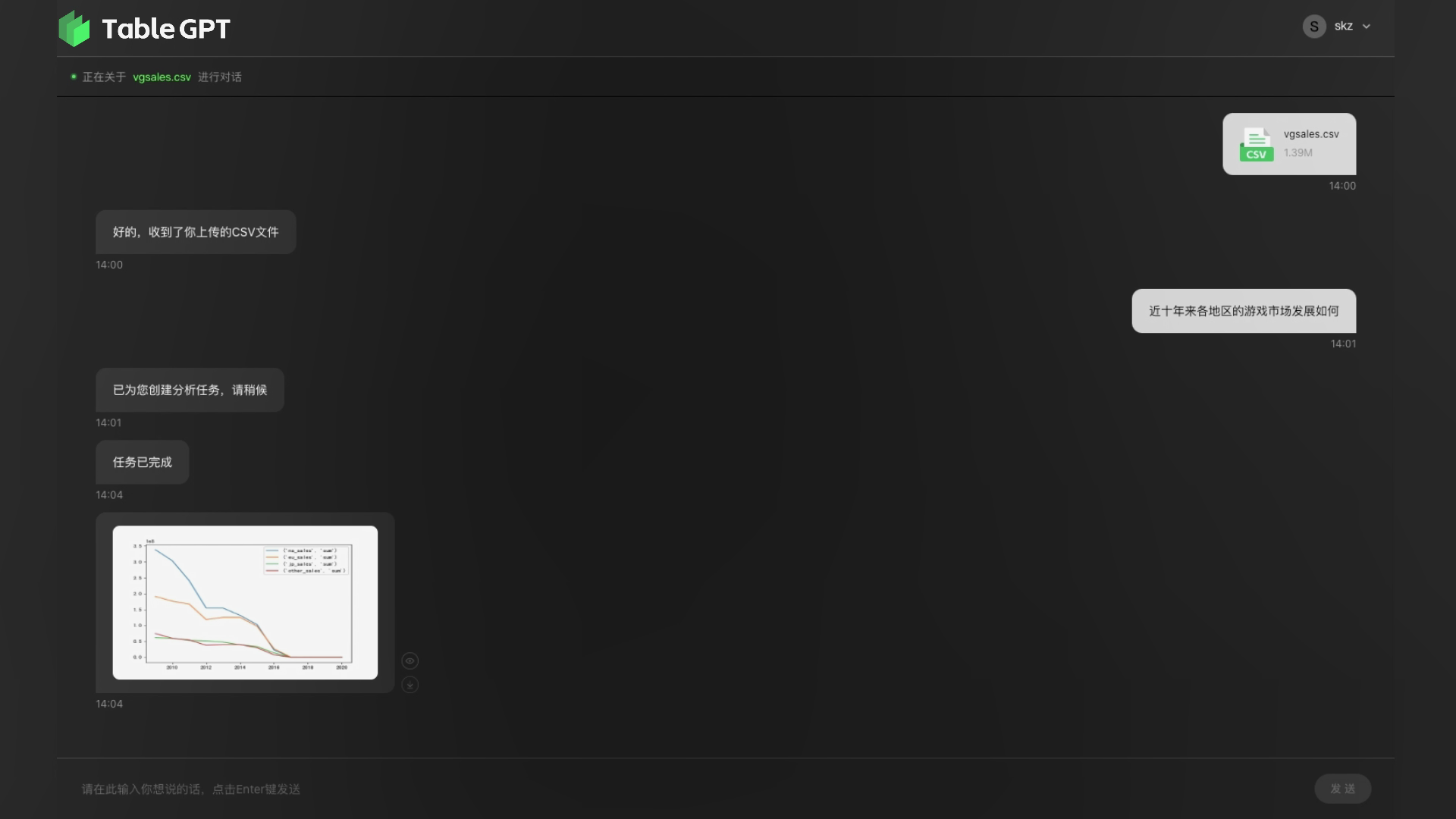}
    \caption{Cases of \ours{}.}
    \label{fig:6}
\end{figure*}

\begin{figure*}[!h]
    \centering
    \includegraphics[scale=0.34]{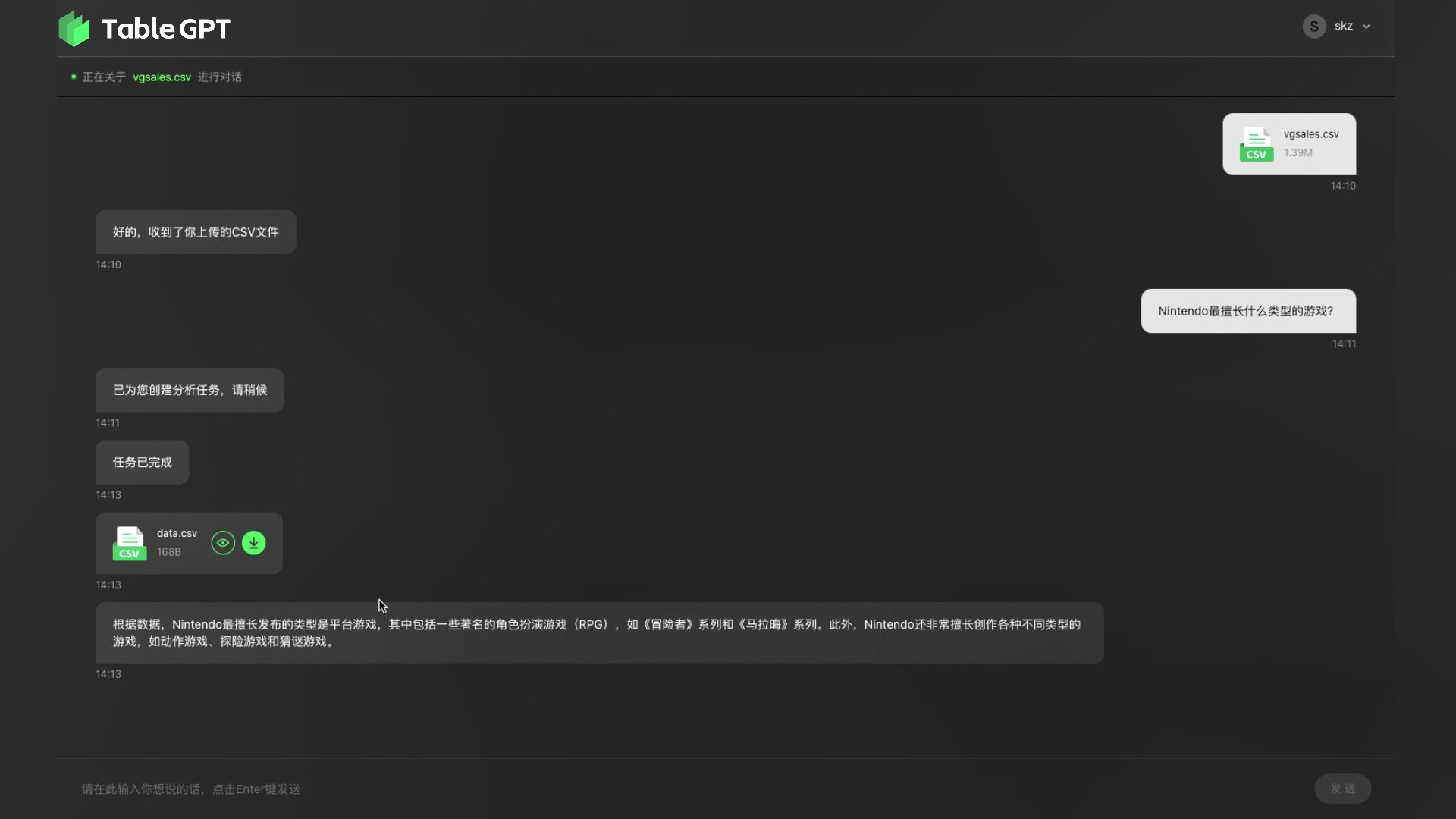}
    \caption{Cases of \ours{}.}
    \label{fig:7}
\end{figure*}

We show some cases in Figure~\ref{fig:1} - \ref{fig:7}. More examples will be released soon.

\section{Conclusion}
We present \ours{}, a large language model designed for table analysis, unifying tables, nature language, and commands. It enables a variety of functions like answering questions, manipulating data, visualizing information, generating analysis reports, and making predictions. 
Technically, \ours{} addresses several major challenges in developing a natural language-driven framework for table data processing, including comprehensive table understanding, instruction chain generation, and domain-specific fine-tuning.
We believe \ours{} has the potential to reshape the landscape of tabular data processing, accelerating the efficiency of table modeling and exploratory data analysis (EDA), and empowering various domains like finance, transportation, scientific research, etc.

\clearpage

\bibliography{neurips_2023}
\bibliographystyle{plain}


\end{document}